\pdfoutput=1

\documentclass[11pt]{article}

\usepackage[]{acl}

\usepackage{times}
\usepackage{latexsym}
\usepackage{booktabs}
\usepackage{bm}
\usepackage{bbm}

\usepackage[T1]{fontenc}

\usepackage[utf8]{inputenc}
\usepackage[hang,flushmargin]{footmisc}

\usepackage{microtype}
\usepackage{amsmath}
\usepackage{tabularx}
\usepackage{graphicx}
\usepackage{multirow}
\usepackage{textcomp}

\title{SimKGC: Simple Contrastive Knowledge Graph Completion with \\ Pre-trained Language Models}

\author{Liang Wang$^1$\thanks{\ \ Work done while at Yuanfudao AI Lab.} \and Wei Zhao$^2$ \and Zhuoyu Wei$^2$ \and Jingming Liu$^2$ \\
        $^1$Microsoft Research Asia \\
        $^2$Yuanfudao AI Lab, Beijing, China \\
    {\tt wangliang@microsoft.com} \\ {\tt \{zhaowei01,weizhuoyu,liujm\}@yuanfudao.com} \\}

\begin{document}
\maketitle
\begin{abstract}
Knowledge graph completion (KGC) aims to reason over known facts
and infer the missing links.
Text-based methods such as KG-BERT ~\citep{yao2019kg}
learn entity representations from natural language descriptions,
and have the potential for inductive KGC.
However,
the performance of text-based methods still largely lag behind
graph embedding-based methods
like TransE ~\citep{bordes2013translating} and RotatE ~\citep{sun2018rotate}.
In this paper,
we identify that the key issue is efficient contrastive learning.
To improve the learning efficiency,
we introduce three types of negatives:
in-batch negatives,
pre-batch negatives,
and self-negatives which act as a simple form of hard negatives.
Combined with InfoNCE loss,
our proposed model SimKGC can substantially outperform embedding-based methods
on several benchmark datasets.
In terms of mean reciprocal rank (MRR),
we advance the state-of-the-art by +19\% on WN18RR,
+6.8\% on the Wikidata5M transductive setting,
and +22\% on the Wikidata5M inductive setting.
Thorough analyses are conducted to gain insights into each component.
Our code is available at ~\url{https://github.com/intfloat/SimKGC}.
\end{abstract}

\section{Introduction}
Large-scale knowledge graphs (KGs) are important components for knowledge-intensive applications,
such as question answering ~\citep{sun-etal-2019-pullnet},
recommender systems ~\citep{Huang2018ImprovingSR},
and intelligent conversational agents ~\citep{Dinan2019WizardOW} etc.
KGs usually consist of a set of triples ($h$, $r$, $t$),
where $h$ is the head entity, $r$ is the relation, and $t$ is the tail entity.
Popular public KGs include Freebase ~\citep{bollacker2008freebase},
Wikidata ~\citep{vrandevcic2014wikidata},
YAGO ~\citep{suchanek2007yago},
ConceptNet ~\citep{speer2017conceptnet},
and WordNet ~\citep{miller1995wordnet} etc.
Despite their usefulness in practice,
they are often incomplete.
Knowledge graph completion (KGC) techniques are necessary for the automatic construction
and verification of knowledge graphs.

\begin{figure}[ht]
\begin{center}
 \includegraphics[width=0.95\linewidth]{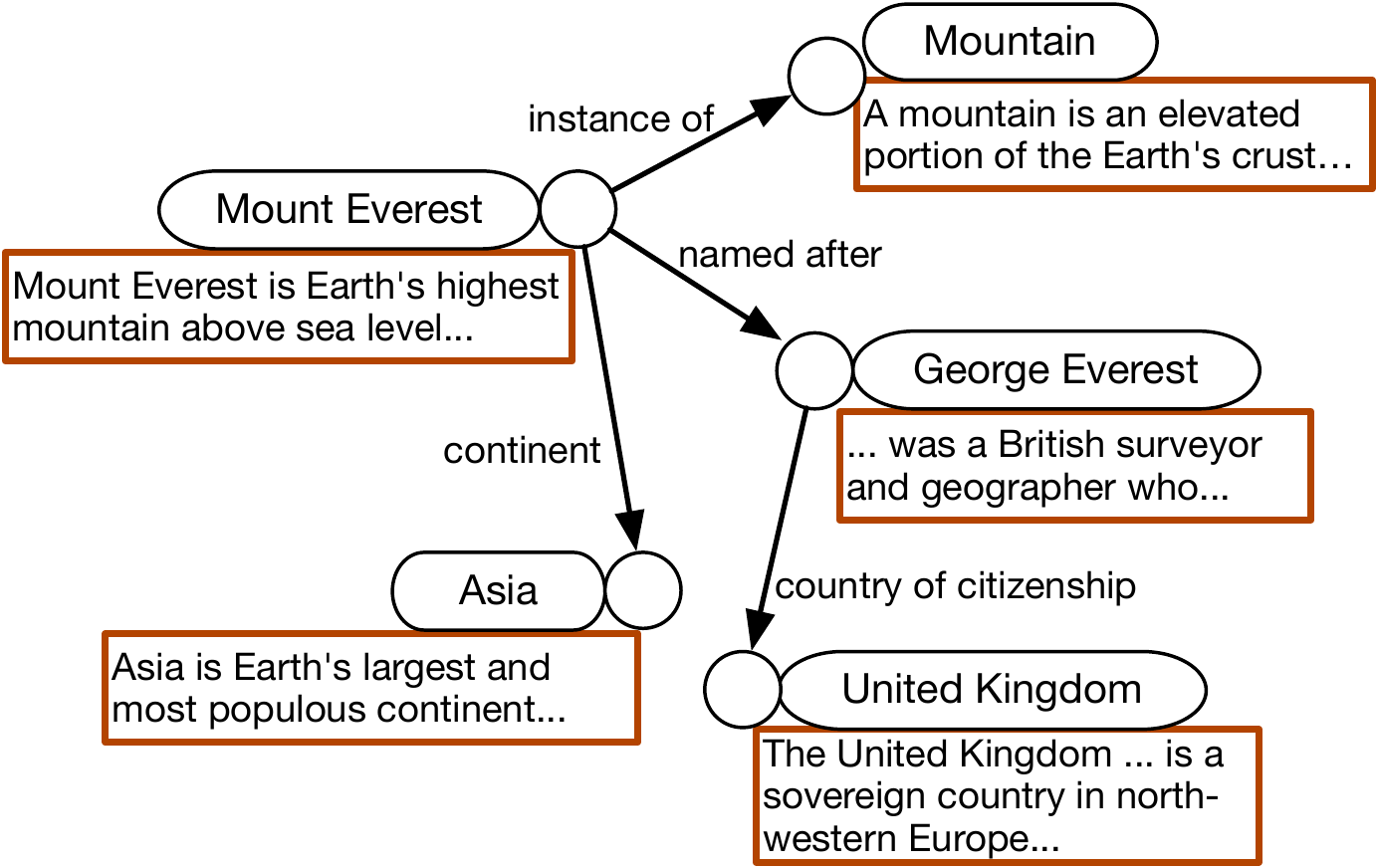}
 \caption{An example of knowledge graph.
 Each entity has its name and textual descriptions.}
 \label{fig:kg}
\end{center}
\end{figure}

Existing KGC methods can be categorized into two families:
embedding-based and text-based methods.
Embedding-based methods map each entity and relation into a low-dimensional vector,
without using any side information such as entity descriptions.
This family includes TransE ~\citep{bordes2013translating}, TransH ~\citep{wang2014knowledge},
RotatE ~\citep{sun2018rotate}, and TuckER ~\citep{balazevic-etal-2019-tucker} etc.
By comparison,
text-based methods ~\citep{yao2019kg,xie2016representation,wang2021kepler}
incorporate available texts for entity representation learning,
as shown in Figure ~\ref{fig:kg}.
Intuitively,
text-based methods should outperform embedding-based counterparts
since they have access to additional input signals.
However,
results on popular benchmarks (e.g., WN18RR, FB15k-237, Wikidata5M)
tell a different story:
text-based methods still lag behind even with pre-trained language models.

We hypothesize that the key issue for such performance degradation
is the inefficiency in contrastive learning.
Embedding-based methods do not involve the expensive computation of text encoders
and thus can be extremely efficient to train with a large negative sample size.
For example,
the default configuration of RotatE ~\footnote{\url{https://github.com/DeepGraphLearning/graphvite}}
trains $1000$ epochs with a negative sample size of $64$ on the Wikidata5M dataset.
While the text-based method KEPLER ~\citep{wang2021kepler}
can only train $30$ epochs with a negative sample size of $1$
due to the high computational cost incurred by RoBERTa.

In this paper,
inspired by the recent progress on contrastive learning,
we introduce three types of negatives to improve the text-based KGC method:
in-batch negatives,
pre-batch negatives,
and self-negatives.
By adopting bi-encoder instead of cross-encoder ~\citep{yao2019kg} architecture,
the number of in-batch negatives can be increased by using a larger batch size.
Vectors from previous batches are cached and act as pre-batch negatives ~\citep{karpukhin-etal-2020-dense}.
Additionally,
mining hard negatives can be beneficial for improving contrastive learning.
We find that the head entity itself can serve as hard negatives,
which we call ``self-negatives''.
As a result,
the negative sample size can be increased to the scale of thousands.
We also propose to change the loss function from margin-based ranking loss to InfoNCE,
which can make the model focus on hard negatives.

One advantage of text-based methods is that
they enable inductive entity representation learning.
Entities that are not seen during training can still be appropriately modeled,
while embedding-based methods like TransE can only reason under the transductive setting
~\footnote{All entities in the test set also appear in the training set.}.
Inductive knowledge graph completion is important in the real world
as new entities are coming out every day.
Moreover,
text-based methods can leverage state-of-the-art pre-trained language models to learn better representations.
A line of recent work ~\citep{shin-etal-2020-autoprompt,petroni-etal-2019-language}
attempts to elicit the implicitly stored knowledge from BERT.
The task of KGC can also be regarded as a way to retrieve such knowledge.

Two entities are more likely to be related
if connected by a short path in the graph.
Empirically,
we find that text-based models heavily rely on the semantic match
and ignore such topological bias to some degree.
We propose a simple re-ranking strategy
by boosting the scores of the head entity's $k$-hop neighbors.

We evaluate our proposed model SimKGC
by conducting experiments on three popular benchmarks:
WN18RR, FB15k-237, and Wikidata5M (both transductive and inductive settings).
According to the automatic evaluation metrics (MRR, Hits@\{1,3,10\}),
SimKGC outperforms state-of-the-art methods by a large margin
on the WN18RR (MRR $47.6 \rightarrow 66.6$),
Wikidata5M transductive setting (MRR $29.0 \rightarrow 35.8$),
and inductive setting (MRR $49.3 \rightarrow 71.4$).
On the FB15k-237 dataset,
our results are also competitive.
To help better understand our proposed method,
we carry out a series of analyses
and report human evaluation results.
Hopefully,
SimKGC will facilitate the future development of better KGC systems.

\section{Related Work}

\noindent
\textbf{Knowledge Graph Completion }
involves modeling multi-relational data
to aid automatic construction of large-scale KGs.
In translation-based methods
such as TransE ~\citep{bordes2013translating} and TransH ~\citep{wang2014knowledge},
a triple ($h$, $r$, $t$) is a relation-specific translation
from the head entity $h$ to tail entity $t$.
Complex number embeddings are introduced by ~\citet{trouillon2016complex}
to increase the model's expressiveness.
RotatE ~\citep{sun2018rotate} models a triple as relational rotation in complex space.
~\citet{nickel2011three,balazevic-etal-2019-tucker} treat KGC as a 3-D binary tensor factorization problem
and investigate the effectiveness of several factorization techniques.
Some methods attempt to incorporate entity descriptions.
DKRL ~\citep{xie2016representation} uses a CNN to encode texts,
while KG-BERT ~\citep{yao2019kg}, StAR ~\citep{wang2021structure}, and BLP ~\citep{daza2021inductive}
both adopt pre-trained language models to compute entity embeddings.
GraIL ~\citep{Teru2020InductiveRP} and BERTRL ~\citep{Zha2021InductiveRP}
conduct inductive relation prediction by utilizing subgraph or path information.
In terms of benchmark performance ~\citep{wang2021kepler},
text-based methods still underperform methods like RotatE.
\newline

\noindent
\textbf{Pre-trained Language Models }
including BERT ~\citep{devlin-etal-2019-bert},
GPT ~\citep{radford2018improving}, and T5 ~\citep{Raffel2020ExploringTL}
have led to a learning paradigm shift in NLP.
Models are first pre-trained on large amounts of unlabeled text corpora
with language modeling objectives,
and then fine-tuned on downstream tasks.
Considering their good performance in few-shot and
even zero-shot scenarios ~\citep{Brown2020LanguageMA},
one interesting question is:
``Can pre-trained language models be
used as knowledge bases?''
~\citet{petroni-etal-2019-language} proposed to probe language models
with manually designed prompts.
A series of following work ~\citep{shin-etal-2020-autoprompt,zhong2021factual,jiang-etal-2020-know}
focus on finding better prompts
to elicit the knowledge implicitly stored in the model parameters.
Another line of work ~\citep{zhang-etal-2019-ernie,liu2020k,wang2021kepler}
injects symbolic knowledge into language model pre-training,
and shows some performance boost on several knowledge-intensive tasks.
\newline

\noindent
\textbf{Contrastive Learning }
learns useful representations by contrasting
between positives and negatives ~\citep{le2020contrastive}.
The definitions of positives and negatives are task-specific.
In self-supervised vision representation learning
~\citep{chen2020simple,he2020momentum,grill2020bootstrap},
a positive pair is two augmented views of the same image,
while a negative pair is two augmented views of different images.
Recently,
contrastive learning paradigm has witnessed great successes
in many different fields,
including multi-modal pre-training ~\citep{radford2021learning},
video-text retrieval ~\citep{Liu2021HiTHT},
and natural language understanding ~\citep{gunel2020supervised} etc.
In the NLP community,
by leveraging the supervision signals from natural language inference data ~\citep{gao2021simcse},
QA pairs ~\citep{ni2021sentence},
and parallel corpora ~\citep{Wang2021AligningCS},
these methods have surpassed non-contrastive methods ~\citep{Reimers2019SentenceBERTSE}
on semantic similarity benchmarks.
~\citet{karpukhin-etal-2020-dense,qu-etal-2021-rocketqa,xiong2020approximate}
adopt contrastive learning to improve dense passage retrieval
for open-domain question answering,
where the positive passages are the ones containing the correct answer.

\section{Methodology}

\subsection{Notations}

A knowledge graph $\mathcal{G}$ is a directed graph,
where the vertices are entities $\mathcal{E}$,
and each edge can be represented as a triple ($h$,$r$,$t$),
where $h$, $r$, and $t$ correspond to
head entity, relation, and tail entity, respectively.
The link prediction task of KGC is to infer the missing triples given an incomplete $\mathcal{G}$.
Under the widely adopted entity ranking evaluation protocol,
tail entity prediction ($h$, $r$, $?$) requires ranking all entities given $h$ and $r$,
similarly for head entity prediction ($?$, $r$, $t$).
In this paper,
for each triple ($h$,$r$,$t$),
we add an inverse triple ($t$,$r^{-1}$,$h$),
where $r^{-1}$ is the inverse relation of $r$.
Based on such reformulation,
we only need to deal with the tail entity prediction problem ~\citep{Malaviya2020CommonsenseKB}.

\subsection{Model Architecture} ~\label{sec:arch}

Our proposed model SimKGC adopts a bi-encoder architecture.
Two encoders are initialized with the same pre-trained language model
but do not share parameters.

Given a triple ($h$,$r$,$t$),
the first encoder BERT$_{hr}$ is used to compute
the relation-aware embedding for the head entity $h$.
We first concatenate the textual descriptions of entity $h$
and relation $r$ with a special symbol [SEP] in between.
BERT$_{hr}$ is applied to get the last-layer hidden states.
Instead of directly using the hidden state of the first token,
we use mean pooling followed by L$_2$ normalization
to get the relation-aware embedding $\mathbf{e}_{hr}$,
as mean pooling has been shown to result in
better sentence embeddings ~\citep{gao2021simcse,Reimers2019SentenceBERTSE}.
$\mathbf{e}_{hr}$ is relation-aware
since different relations will have different inputs
and thus have different embeddings,
even though the head entity is the same.

Similarly,
the second encoder BERT$_{t}$ is used to
compute the L$_2$-normalized embedding $\mathbf{e}_t$ for the tail entity $t$.
The input for BERT$_{t}$ only consists of the textual description for entity $t$.

Since the embeddings $\mathbf{e}_{hr}$ and $\mathbf{e}_{t}$ are both L$_2$ normalized,
the cosine similarity $\cos(\mathbf{e}_{hr}, \mathbf{e}_{t})$
is simply the dot product between two embeddings:
\begin{equation} \label{eq:cos}
\cos(\mathbf{e}_{hr}, \mathbf{e}_{t}) =
\frac{\mathbf{e}_{hr}\cdot\mathbf{e}_t}{\lVert \mathbf{e}_{hr} \rVert \lVert \mathbf{e}_t \rVert}
=\mathbf{e}_{hr}\cdot\mathbf{e}_t
\end{equation}

For tail entity prediction ($h$, $r$, $?$),
we compute the cosine similarity between $\mathbf{e}_{hr}$
and all entities in $\mathcal{E}$,
and predict the one with the largest score:
\begin{equation} \label{eq:argmax}
\underset{t_i}{\mathrm{argmax}}\ \cos(\mathbf{e}_{hr}, \mathbf{e}_{t_i}),\ t_i \in \mathcal{E}
\end{equation}

\subsection{Negative Sampling}

For knowledge graph completion,
the training data only consists of positive triples.
Given a positive triple ($h$, $r$, $t$),
``negative sampling'' needs to sample one or more negative triples
to train discriminative models.
Most existing methods randomly corrupt $h$ or $t$
and then filter out false negatives that appear in the training graph $\mathcal{G}$.
The negatives for different triples are not shared and therefore independent.
The typical number of negatives are $\sim64$ for embedding-based methods ~\citep{sun2018rotate},
and $\sim5$ for text-based methods ~\citep{wang2021structure}.
We combine three types of negatives
to improve the training efficiency
without incurring significant computational and memory overhead.
\newline

\noindent
\textbf{In-batch Negatives (IB)}
This is a widely adopted strategy in visual representation learning ~\citep{chen2020simple}
and dense passage retrieval ~\citep{karpukhin-etal-2020-dense} etc.
Entities within the same batch can be used as negatives.
Such in-batch negatives allow the efficient reuse of entity embeddings for bi-encoder models.
\newline

\noindent
\textbf{Pre-batch Negatives (PB)}
The disadvantage of in-batch negatives is that
the number of negatives is coupled with batch size.
Pre-batch negatives ~\citep{Lee2021LearningDR}
use entity embeddings from previous batches.
Since these embeddings are computed with an earlier version of model parameters,
they are not consistent with in-batch negatives.
Usually,
only $1$ or $2$ pre-batches are used.
Other methods like MoCo ~\citep{he2020momentum} can also provide more negatives.
We leave the investigation of MoCo as future work.
\newline

\noindent
\textbf{Self-Negatives (SN)}
Besides increasing the number of negatives,
mining hard negatives ~\citep{gao2021simcse,xiong2020approximate} is also important
for improving contrastive representation learning.
For tail entity prediction ($h$, $r$, $?$),
text-based methods
tend to assign a high score to the head entity $h$,
likely due to the high text overlap.
To mitigate this issue,
we propose self-negatives that use the head entity $h$ as hard negatives.
Including self-negatives can make the model rely less
on the spurious text match.
\newline

We use $\mathcal{N}_\text{IB}$, $\mathcal{N}_\text{PB}$, and $\mathcal{N}_\text{SN}$
to denote the aforementioned three types of negatives.
During training,
there may exist some false negatives.
For example,
the correct entity happens to appear in another triple within the same batch.
We filter out such entities with a binary mask
~\footnote{False negatives that do not appear in the training data will not be filtered.}.
Combining them all,
the collection of negatives $\mathcal{N}(h,r)$ is:
\begin{equation}
\{t'| t' \in \mathcal{N}_\text{IB} \cup \mathcal{N}_\text{PB} \cup \mathcal{N}_\text{SN},
\text{(}h,r,t'\text{)} \notin \mathcal{G}\}
\end{equation}

Assume the batch size is $1024$,
and $2$ pre-batches are used,
we would have $|\mathcal{N}_\text{IB}| = 1024 - 1$,
$|\mathcal{N}_\text{PB}| = 2 \times 1024$,
$|\mathcal{N}_\text{SN}| = 1$,
and $|\mathcal{N}(h,r)| = 3072$ negatives in total.

\subsection{Graph-based Re-ranking}

Knowledge graphs often exhibit spatial locality.
Nearby entities are more likely to be related than entities that are far apart.
Text-based KGC methods are good at capturing semantic relatedness
but may not fully capture such inductive bias.
We propose a simple graph-based re-ranking strategy:
increase the score of candidate tail entity $t_i$
by $\alpha \ge 0$ if $t_i$ is in $k$-hop neighbors $\mathcal{E}_k(h)$
of the head entity $h$ based on the graph from training set:
\begin{equation} \label{eq:rerank}
\underset{t_i}{\mathrm{argmax}}\ \cos(\mathbf{e}_{hr},\mathbf{e}_{t_i}) + \alpha\mathbbm{1}(t_i \in \mathcal{E}_k(h))
\end{equation}

\subsection{Training and Inference}
During training,
we use InfoNCE loss with additive margin ~\citep{chen2020simple,yang2019improving}:
\begin{equation} \label{eq:infonce}
    \mathcal{L} = -\log \frac{e^{(\phi(h,r,t)-\gamma)/\tau}}
    {e^{(\phi(h,r,t)-\gamma)/\tau} + \sum_{i=1}^{|\mathcal{N}|}{e^{\phi(h,r,t_i')/\tau}}}
\end{equation}

The additive margin $\gamma > 0$ encourages the model
to increase the score of the correct triple ($h$,$r$,$t$).
$\phi(h,r,t)$ is the score function for a candidate triple,
here we define $\phi(h,r,t)=\cos(\mathbf{e}_{hr}, \mathbf{e}_{t}) \in [-1,1]$
as in Equation ~\ref{eq:cos}.
The temperature $\tau$ can adjust the relative importance of negatives,
smaller $\tau$ makes the loss put more emphasis on hard negatives,
but also risks over-fitting label noise.
To avoid tuning $\tau$ as a hyperparameter,
we re-parameterize $\log \frac{1}{\tau}$ as a learnable parameter.

For inference,
the most time-consuming part is $O(|\mathcal{E}|)$ BERT forward pass computation of entity embeddings.
Assume there are $|\mathcal{T}|$ test triples.
For each triple ($h$, $r$, $?$) and ($t$, $r^{-1}$, $?$),
we need to compute the relation-aware head entity embedding
and use a dot product to get the ranking score for all entities.
In total,
SimKGC needs $|\mathcal{E}|+2\times|\mathcal{T}|$ BERT forward passes,
while cross-encoder models like KG-BERT ~\citep{yao2019kg}
needs $|\mathcal{E}|\times2\times|\mathcal{T}|$.
Being able to scale to large datasets is important for practical usage.
For bi-encoder models,
we can pre-compute the entity embeddings
and retrieve top-k entities efficiently
with the help of fast similarity search tools like Faiss ~\citep{Johnson2021BillionScaleSS}.

\section{Experiments}

\begin{table*}[ht]
\centering
\begin{tabular}{@{}l|lllll@{}}
\toprule
dataset          & \#entity & \#relation & \#train & \#valid & \#test \\ \midrule
WN18RR           &   $40,943$  &    $11$    &  $86,835$   &   $3034$   &  $3134$     \\
FB15k-237        &   $14,541$  &    $237$  &  $272,115$  &  $17,535$ & $20,466$ \\
Wikidata5M-Trans &   $4,594,485$   &  $822$   &  $20,614,279$    &  $5,163$  &  $5,163$    \\
Wikidata5M-Ind   &   $4,579,609$   &   $822$ &  $20,496,514$  & $6,699$  &  $6,894$  \\ \bottomrule
\end{tabular}
\caption{Statistics of the datasets used in this paper.
``Wikidata5M-Trans'' and ``Wikidata5M-Ind'' refer to the transductive and inductive settings,
respectively.}
\label{tab:dataset}
\end{table*}

\subsection{Experimental Setup}

\noindent
\textbf{Datasets }
We use three datasets for evaluation:
WN18RR,
FB15k-237,
and Wikidata5M ~\citep{wang2021kepler}.
The statistics are shown in Table ~\ref{tab:dataset}.
~\citet{bordes2013translating} proposed the WN18 and FB15k datasets.
Later work ~\citep{toutanova-etal-2015-representing,dettmers2018convolutional}
showed that these two datasets suffer from test set leakage
and released WN18RR and FB15k-237 datasets by removing the inverse relations.
The WN18RR dataset consists of $\sim41k$ synsets and $11$ relations from WordNet ~\citep{miller1995wordnet},
and the FB15k-237 dataset consists of $\sim15k$ entities and $237$ relations from Freebase.
The Wikidata5M dataset is much larger in scale with $\sim5$ million entities and $\sim20$ million triples.
It provides two settings: transductive and inductive.
For the transductive setting,
all entities in the test set also appear in the training set,
while for the inductive setting,
there is no entity overlap between train and test set.
We use ``Wikidata5M-Trans'' and ``Wikidata5M-Ind'' to indicate these two settings.

For textual descriptions,
we use the data provided by KG-BERT ~\citep{yao2019kg} for WN18RR and FB15k-237 datasets.
The Wikidata5M dataset already contains descriptions for all entities and relations.
\newline

\noindent
\textbf{Evaluation Metrics }
Following previous work,
our proposed KGC model is evaluated with entity ranking task:
for each test triple $(h,r,t)$,
tail entity prediction ranks all entities to predict $t$ given $h$ and $r$,
similarly for head entity prediction.
We use four automatic evaluation metrics:
mean reciprocal rank (MRR),
and Hits@$k$($k\in$\{$1$,$3$,$10$\}) (H@$k$ for short).
MRR is the average reciprocal rank of all test triples.
H@$k$ calculates the proportion of correct entities ranked among the top-$k$.
MRR and H@$k$ are reported under the \emph{filtered setting} ~\citep{bordes2013translating},
The \emph{filtered setting} ignores the scores of all known true triples
in the training, validation, and test set.
All metrics are computed by averaging over two directions:
head entity prediction and tail entity prediction.

We also conduct a human evaluation on the Wikidata5M dataset
to provide a more accurate estimate of the model's performance.
\newline

\noindent
\textbf{Hyperparameters }
The encoders are initialized with \emph{bert-base-uncased} (English).
Using better pre-trained language models is expected to improve performance further.
Most hyperparameters except learning rate and training epochs
are shared across all datasets
to avoid dataset-specific tuning.
We conduct grid search on learning rate with ranges \{$10^{-5}$, $3\times10^{-5}$, $5\times10^{-5}$\}.
Entity descriptions are truncated to a maximum of $50$ tokens.
Temperature $\tau$ is initialized to $0.05$,
and the additive margin for InfoNCE loss is $0.02$.
For re-ranking,
we set $\alpha = 0.05$.
$2$ pre-batches are used with logit weight $0.5$.
We use AdamW optimizer with linear learning rate decay.
Models are trained with batch size $1024$ on $4$ V100 GPUs.
For the WN18RR, FB15k-237, and Wikidata5M (both settings) datasets,
we train for $50$, $10$, and $1$ epochs, respectively.
Please see Appendix ~\ref{app:setup} for more details.

\subsection{Main Results}

\begin{table*}[ht]
\centering
\scalebox{0.9}{\begin{tabular}{l|cccc|cccc}
\hline
\multirow{2}{*}{Method} & \multicolumn{4}{c|}{Wikidata5M-Trans} & \multicolumn{4}{c}{Wikidata5M-Ind}  \\ \cline{2-9}
                        & MRR & H@1 & H@3 & H@10 & MRR & H@1 & H@3 & H@10 \\ \hline
\multicolumn{9}{l}{\textit{embedding-based methods}}         \\ \hline
TransE ~\citep{bordes2013translating} &  25.3 & 17.0 & 31.1 & 39.2  &  - & - & - &  - \\
RotatE ~\citep{sun2018rotate} & 29.0 & 23.4 & 32.2 & 39.0 &  - & - & - & -  \\ \hline
\multicolumn{9}{l}{\textit{text-based methods}}         \\ \hline
DKRL ~\citep{xie2016representation} & 16.0 & 12.0 & 18.1 & 22.9 & 23.1 & 5.9 & 32.0 & 54.6 \\
KEPLER ~\citep{wang2021kepler} & 21.0 & 17.3 & 22.4 & 27.7 & 40.2 & 22.2 & 51.4 & 73.0 \\
BLP-ComplEx ~\citep{daza2021inductive} & - & - & - & - &  48.9 & 26.2 & 66.4 & 87.7 \\
BLP-SimplE ~\citep{daza2021inductive} & - & - & - & - &  49.3 & 28.9 & 63.9 & 86.6 \\ \hline \hline
SimKGC$_\text{IB}$  & 35.3 & 30.1 & 37.4 & \textbf{44.8} & 60.3 & 39.5 & 77.8 & 92.3 \\
SimKGC$_\text{IB+PB}$ & 35.4 & 30.2 & 37.3 & \textbf{44.8} & 60.2 & 39.4 & 77.7 & \textbf{92.4} \\
SimKGC$_\text{IB+SN}$ & 35.6 & 31.0 & 37.3 & 43.9 & 71.3 & 60.7 & \textbf{78.7} & 91.3 \\
SimKGC$_\text{IB+PB+SN}$ & \textbf{35.8} & \textbf{31.3} & \textbf{37.6} & 44.1 & \textbf{71.4} & \textbf{60.9} & 78.5 & 91.7 \\ \hline
\end{tabular}}
\caption{Main results for the Wikidata5M dataset.
``IB'', ``PB'', and ``SN'' refer to in-batch negatives,
pre-batch negatives, and self-negatives respectively.
Embedding-based methods are inherently unable to perform inductive KGC.
According to the evaluation protocol by ~\citet{wang2021kepler},
the inductive setting only ranks $7,475$ entities in the test set,
while the transductive setting ranks $\sim4.6$ million entities,
so the reported metrics for the inductive setting are much higher.
Results are statistically significant under paired student's t-test with p-value $0.05$.}
\label{tab:wikidata}
\end{table*}

\begin{table*}[ht]
\centering
\scalebox{0.9}{\begin{tabular}{l|cccc|cccc}
\hline
\multirow{2}{*}{Method} & \multicolumn{4}{c|}{WN18RR}      & \multicolumn{4}{c}{FB15k-237}  \\ \cline{2-9}
                        & MRR & H@1 & H@3 & H@10 & MRR & H@1 & H@3 & H@10 \\ \hline
\multicolumn{9}{l}{\textit{embedding-based methods}}         \\ \hline
TransE ~\citep{bordes2013translating}$^\dagger$ &  24.3 & 4.3  & 44.1 & 53.2 & 27.9 & 19.8 & 37.6 & 44.1 \\
DistMult ~\citep{yang2014embedding}$^\dagger$   &  44.4 & 41.2 & 47.0 & 50.4  &  28.1 & 19.9 & 30.1 & 44.6 \\
RotatE ~\citep{sun2018rotate}$^\dagger$  & 47.6 & 42.8 & 49.2 & 57.1 & 33.8 & 24.1 &  37.5 & 53.3 \\
TuckER ~\citep{balazevic-etal-2019-tucker}$^\dagger$ & 47.0 & 44.3 & 48.2 & 52.6 & \textbf{35.8} & \textbf{26.6} & \textbf{39.4} & \textbf{54.4} \\ \hline
\multicolumn{9}{l}{\textit{text-based methods}}         \\ \hline
KG-BERT ~\citep{yao2019kg} & 21.6 & 4.1 &  30.2 & 52.4 &  -  &  - &  - &  42.0 \\
MTL-KGC ~\citep{kim-etal-2020-multi} & 33.1 & 20.3 & 38.3 & 59.7 & 26.7 & 17.2 & 29.8 & 45.8 \\
StAR ~\citep{wang2021structure} &  40.1 & 24.3 &  49.1 & 70.9 &  29.6 & 20.5 & 32.2 & 48.2 \\ \hline \hline
SimKGC$_\text{IB}$  & \textbf{67.1} & 58.5 & \textbf{73.1} & \textbf{81.7} & 33.3 & 24.6 & 36.2  & 51.0 \\
SimKGC$_\text{IB+PB}$ & 66.6 &  57.8  & 72.3 & \textbf{81.7} &  33.4 & 24.6 & \textbf{36.5} & \textbf{51.1} \\
SimKGC$_\text{IB+SN}$ & 66.7 & \textbf{58.8} & 72.1 & 80.5 & 33.4 & 24.7 & 36.3 & 50.9 \\
SimKGC$_\text{IB+PB+SN}$ & 66.6 & 58.7 & 71.7 & 80.0 & \textbf{33.6} & \textbf{24.9} & 36.2 & \textbf{51.1} \\ \hline
\end{tabular}}
\caption{Main results for WN18RR and FB15k-237 datasets.
$^\dagger$: numbers are from ~\citet{wang2021structure}.}
\label{tab:wn_and_fb}
\end{table*}

We reuse the numbers reported by ~\citet{wang2021kepler} for TransE and DKRL,
and the results for RotatE are from the official GraphVite
~\footnote{\url{https://graphvite.io/docs/latest/benchmark}} benchmark.
In Table ~\ref{tab:wikidata} and ~\ref{tab:wn_and_fb},
our proposed model SimKGC$_\text{IB+PB+SN}$
outperforms state-of-the-art methods by a large margin
on the WN18RR,
Wikidata5M-Trans,
and Wikidata5M-Ind datasets,
but slightly lags behind on the FB15k-237 dataset (MRR $33.6\%\ \text{vs}\ 35.8\%$).
To the best of our knowledge,
SimKGC is the first text-based KGC method that
achieves better results than embedding-based counterparts.

We report results for various combinations of negatives.
With in-batch negatives only,
the performance of SimKGC$_\text{IB}$ is already quite strong
thanks to the large batch size (1024) we use.
Adding self-negatives tends to improve H@1 but hurt H@10.
We hypothesize that self-negatives make the model rely less on simple text match.
Thus they have negative impacts on metrics that emphasize recall,
such as H@10.
Combining all three types of negatives generally has the best results but not always.

Compared to other datasets,
the graph for the FB15k-237 dataset
is much denser (average degree is $\sim37$ per entity),
and contains fewer entities ($\sim15k$).
To perform well,
models need to learn generalizable inference rules
instead of just modeling textual relatedness.
Embedding-based methods are likely to hold an advantage for this scenario.
It is possible to ensemble our method with embedding-based ones,
as done by ~\citet{wang2021structure}.
Since this is not the main focus of this paper,
we leave it as future work.
Also,
~\citet{cao-etal-2021-missing} points out that
many links in the FB15k-237 dataset are not predictable based on the available information.
These two reasons help explain the unsatisfactory performance of SimKGC.

Adding self-negatives is particularly helpful for the inductive setting of Wikidata5M dataset,
with MRR rising from $60.3\%$ to $71.3\%$.
For inductive KGC,
text-based models rely more heavily on text match than the transductive setting.
Self negatives can prevent the model from simply predicting the given head entity.

In terms of inference time,
the most expensive part is the forward pass with BERT.
For the Wikidata5M-Trans dataset,
SimKGC requires $\sim40$ minutes to compute $\sim4.6$ million embeddings with 2 GPUs,
while cross-encoder models such as KG-BERT ~\citep{yao2019kg}
would require an estimated time of $3000$ hours.
We are not the first work that enables fast inference,
models such as ConvE ~\citep{dettmers2018convolutional} and StAR ~\citep{wang2021structure}
also share similar advantages.
Here we just want to re-emphasize the importance of inference efficiency and scalability
when designing new models.

\section{Analysis}
We conduct a series of analyses to gain further insights
into our proposed model and the KGC task.

\subsection{What Makes SimKGC Excel?}

Compared to existing text-based methods,
SimKGC makes two major changes:
using more negatives,
and switching from margin-based ranking loss to InfoNCE loss.
To guide the future work on knowledge graph completion,
it is crucial to understand which factor contributes most
to the superior performance of SimKGC.

\begin{table}[ht]
\centering
\scalebox{0.88}{\begin{tabular}{lc|cccc}
\hline
loss & \# of neg & MRR & H@1 & H@3 & H@10 \\ \hline
InfoNCE & 255 & \textbf{64.4} & \textbf{53.8} & \textbf{71.7} & \textbf{82.8} \\
InfoNCE & 5 & 48.8 & 31.9 & 60.2 & 80.3 \\
margin & 255 & 39.5 & 28.5 & 44.4 & 61.2 \\
margin & 5 & 38.0 & 27.5 & 42.8 & 58.7 \\ \hline
margin-$\tau$ & 255 & 57.8 & 48.5 & 63.7 & 74.9 \\ \hline
\end{tabular}}
\caption{Analysis of loss function and the number of negatives on the WN18RR dataset.}
\label{tab:ablation_key_factor}
\end{table}

\begin{table*}[ht]
\centering
\scalebox{0.9}{\begin{tabular}{l|l}
\hline
triple   &   (Rest Plaus Historic District, is located in, \textbf{New York}) \\
evidence & \ldots a national historic district located at Marbletown in Ulster County, New York\ldots \\
SimKGC &  Marbletown  \\ \hline
triple   &   (\textbf{Timothy P. Green}, place of birth, St. Louis) \\
evidence & William Douglas Guthrie (born January 17, 1967 in St. Louis, MO) is a professional boxer\ldots \\
SimKGC &  William Douglas Guthrie \\ \hline
triple   &   (TLS termination proxy, instance of, \textbf{networked software}) \\
evidence & \ldots a proxy server that is used by an institution to handle incoming TLS connections\ldots \\
SimKGC &  http server  \\ \hline
triple   &   (1997 IBF World Championships, followed by, \textbf{1999 IBF World Championships}) \\
evidence & \begin{tabular}[c]{@{}l@{}} The 10th IBF World Championships (Badminton) were held in Glasgow, Scotland, \\ between 24 May and 1 June 1997\ldots \end{tabular} \\
SimKGC &  2000 IBF World Junior Championships  \\ \hline
\end{tabular}}
\caption{Examples of SimKGC prediction results on the test set of the Wikidata5M-Trans dataset.
The entity to predict is in bold font.
We only show a snippet of relevant texts in the row of ``evidence'' for space reason.}
\label{tab:case_analysis}
\end{table*}

In Table ~\ref{tab:ablation_key_factor},
we use SimKGC$_\text{IB}$ with batch size $256$ as a baseline.
By reducing the number of negatives from $255$ to $5$,
MRR drops from $64.4$ to $48.8$.
Changing the loss function from InfoNCE to the following margin loss makes MRR drop to $39.5$:
\begin{equation} \label{eq:margin}
    \frac{1}{|\mathcal{N}|}\sum_{i=1}^{|\mathcal{N}|}\max(0, \lambda + \phi(h,r,t'_i) - \phi(h,r,t))
\end{equation}
Consistent with Equation ~\ref{eq:infonce},
$\phi(h,r,t'_i)$ is cosine similarity score for a candidate triple,
and $\lambda = 0.8$.

To summarize,
both InfoNCE loss and a large number of negatives are important factors,
while the loss function seems to have bigger impacts.
For InfoNCE loss,
the hard negatives naturally contribute larger gradients,
and adding more negatives can lead to more robust representations.
~\citet{Wang2021UnderstandingTB} also draws a similar conclusion:
such hardness-aware property is vital for the success of contrastive loss.

We also propose a variant ``margin-$\tau$'' loss
by changing the weight in Equation ~\ref{eq:margin} from $\frac{1}{|\mathcal{N}|}$
to $\frac{\exp(s(t'_i)/\tau)}{\sum_{j=1}^{|\mathcal{N}|}\exp(s(t'_j)/\tau)}$,
where $s(t'_i) = \max(0, \lambda + \phi(h,r,t'_i) - \phi(h,r,t))$
and $\tau = 0.05$.
Similar to InfoNCE loss,
``margin-$\tau$'' loss makes the model pay more attention to hard negatives
and leads to better performance as shown in Table ~\ref{tab:ablation_key_factor}.
It is similar to the ``self-adversarial negative sampling'' proposed by ~\citet{sun2018rotate}.
Most hyperparameters are tuned based on InfoNCE loss.
We expect the margin-$\tau$ loss to achieve better results with a bit more hyperparameter optimization.

\begin{figure}[ht]
\begin{center}
 \includegraphics[width=0.95\linewidth]{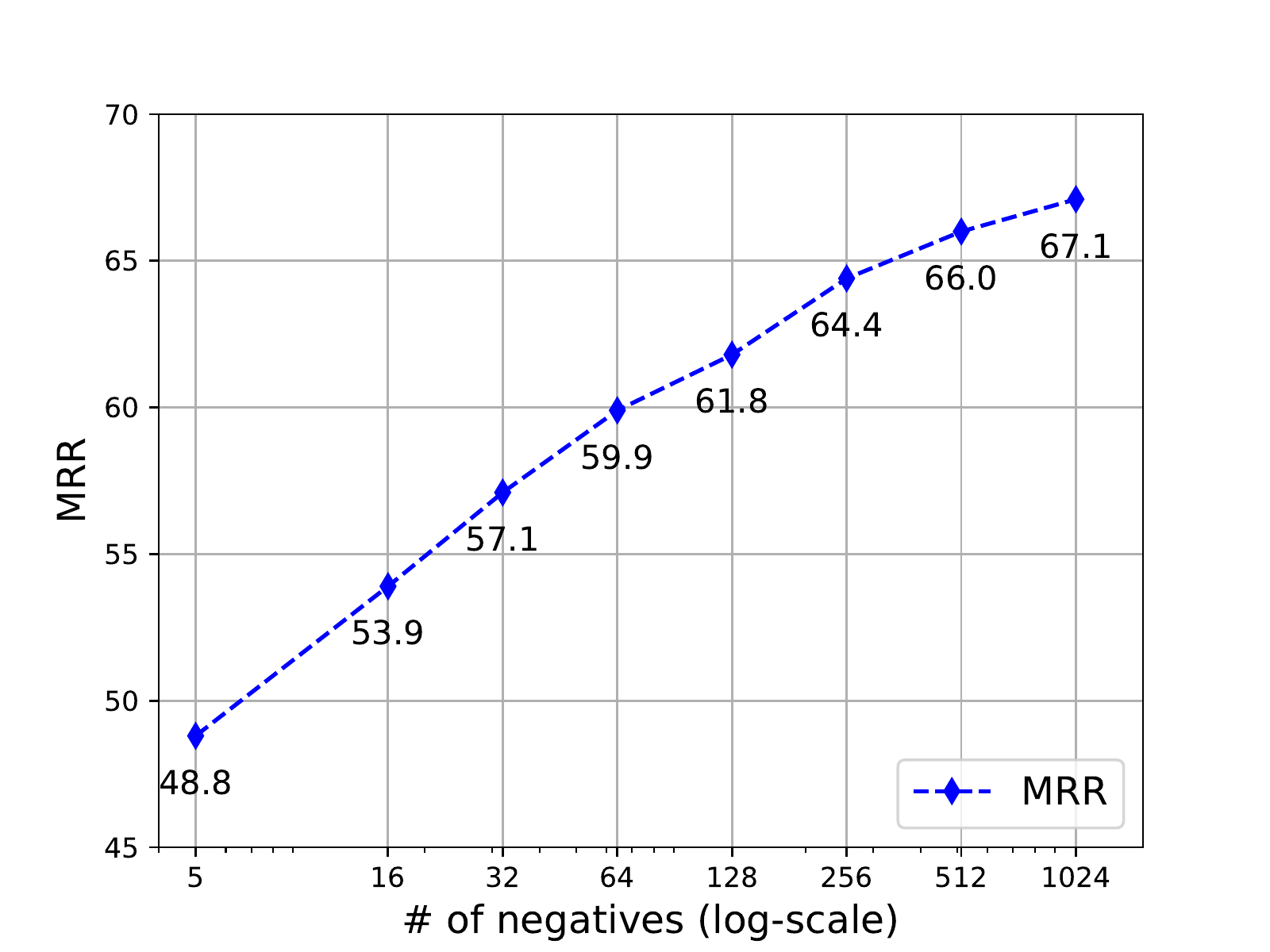}
 \caption{MRR on the WN18RR dataset w.r.t the number of negatives with SimKGC$_\text{IB}$.
 We use a batch size of $1024$ for all experiments,
 and change the number of negatives with a binary mask over the softmax logits.}
 \label{fig:neg_size}
\end{center}
\end{figure}

In Figure ~\ref{fig:neg_size},
we quantitatively illustrate how MRR changes as more negatives are added.
There is a clear trend that
the performance steadily improves from $48.8$ to $67.1$.
However,
adding more negatives requires more GPU memory
and may cause optimization difficulties ~\citep{You2020LargeBO,chen2020simple}.
We do not experiment with batch size larger than $1024$.

\subsection{Ablation on Re-ranking}

\begin{table}[ht]
\centering
\begin{tabular}{c|llll}
\hline
 & MRR & H@1 & H@3 & H@10 \\ \hline
w/ re-rank &  \textbf{35.8} & \textbf{31.3} & \textbf{37.6} & \textbf{44.1} \\
w/o re-rank & 35.5 & 31.0 & 37.3 & 43.9 \\ \hline
\end{tabular}
\caption{Ablation of re-ranking on the Wikidata5M-Trans dataset.}
\label{tab:re-rank}
\end{table}

Our proposed re-ranking strategy is a simple way
to incorporate topological information in the knowledge graph.
For graphs whose connectivity patterns exhibit spatial locality,
re-ranking is likely to help.
In Table ~\ref{tab:re-rank},
we see a slight but stable increase for all metrics on the Wikidata5M-Trans dataset.
Note that this re-ranking strategy does not apply
to inductive KGC
since entities in the test set never appear in the training data.
Exploring more effective ways such as graph neural networks ~\citep{Wu2019ACS}
instead of simple re-ranking would be a future direction.

\subsection{Fine-grained Analysis}

\begin{table}[ht]
\centering
\begin{tabular}{cccc}
\hline
1-1   & 1-n \\ \hline
\begin{tabular}[c]{@{}c@{}}spouse\\ capital of \\lake inflows \\ head of government \end{tabular}
& \begin{tabular}[c]{@{}c@{}}child\\ has part \\ notable work \\ side effect \end{tabular} \\ \hline
n-1  & n-n   \\ \hline
 \begin{tabular}[c]{@{}c@{}} instance of \\ place of birth \\ given name \\ work location \end{tabular}
& \begin{tabular}[c]{@{}c@{}} cast member \\ member of \\ influenced by \\ nominated for \end{tabular} \\ \hline
\end{tabular}
\caption{Examples for different categories of relations on the Wikidata5M-Trans dataset.}
\label{tab:relation_examples}
\end{table}

\begin{table}[ht]
\centering
\begin{tabular}{l|cccc}
\hline
Dataset & 1-1 & 1-n & n-1 & n-n \\ \hline
Wikidata5M-Trans & 30.4 & 8.3 & 71.1 & 10.6 \\
Wikidata5M-Ind & 83.5 & 71.1 & 80.0 & 54.7 \\ \hline
\end{tabular}
\caption{MRR for different kinds of relations
on the Wikidata5M dataset with SimKGC$_\text{IB+PB+SN}$.}
\label{tab:perf_by_relation}
\end{table}

We classify all relations into four categories
based on the cardinality of head and tail arguments
following the rules by ~\citet{bordes2013translating}:
one-to-one(1-1),
one-to-many(1-n),
many-to-one(n-1),
and many-to-many(n-n).
Examples are shown in Table ~\ref{tab:relation_examples}.
As shown in Table ~\ref{tab:perf_by_relation},
predicting the ``n'' side is generally more difficult,
since there are many seemingly plausible answers
that would confuse the model.
Another main reason is the incompleteness of the knowledge graph.
Some predicted triples might be correct based on human evaluation,
especially for 1-n relations in head entity prediction,
such as ``instance of'', ``place of birth'' etc.

In Table ~\ref{tab:case_analysis},
for the first example,
``Marbletown'', ``Ulster County'', and ``New York'' are both correct answers.
The second example illustrates the case for relation ``place of birth'':
a lot of people share the same place of birth,
and some triples may not exist in the knowledge graph.
This helps explain the low performance of ``1-n'' relations for the Wikidata5M-Trans dataset.
In the third example,
SimKGC predicts a closely related but incorrect entity ``http server''.

\subsection{Human Evaluation}

\begin{table}[ht]
\centering
\begin{tabular}{l|lll}
\hline
         & correct & wrong & unknown \\ \hline
($h$, $r$, $?$)  & 24\% & 54\% & 22\% \\
($?$, $r$, $t$) & 74\% & 14\% & 12\% \\
Avg   & 49\%  & 34\% & 17\% \\ \hline
\end{tabular}
\caption{Human evaluation results on the Wikidata5M-Trans dataset.
($h$, $r$, $?$) and ($?$, $r$, $t$) denote tail entity and head entity prediction respectively.
We randomly sample $100$ wrong predictions according to H@1 from test set.
The ``unknown'' category indicates annotators are unable to decide
whether the prediction is correct or wrong
based on the textual information.}
\label{tab:human_eval}
\end{table}

The analyses above suggest that
automatic evaluation metrics such as MRR tend to underestimate the model's performance.
To have a more accurate estimation of the performance,
we conduct human evaluation and list the results in Table ~\ref{tab:human_eval}.
An average of $49\%$ of the wrong predictions according to H@1 are correct according to human annotators.
If we take this into account,
the H@1 of our proposed model would be much higher.
How to accurately measure the performance of KGC systems
is also an interesting future research direction.

\subsection{Entity Visualization}

\begin{figure}[ht]
\begin{center}
 \includegraphics[width=0.95\linewidth]{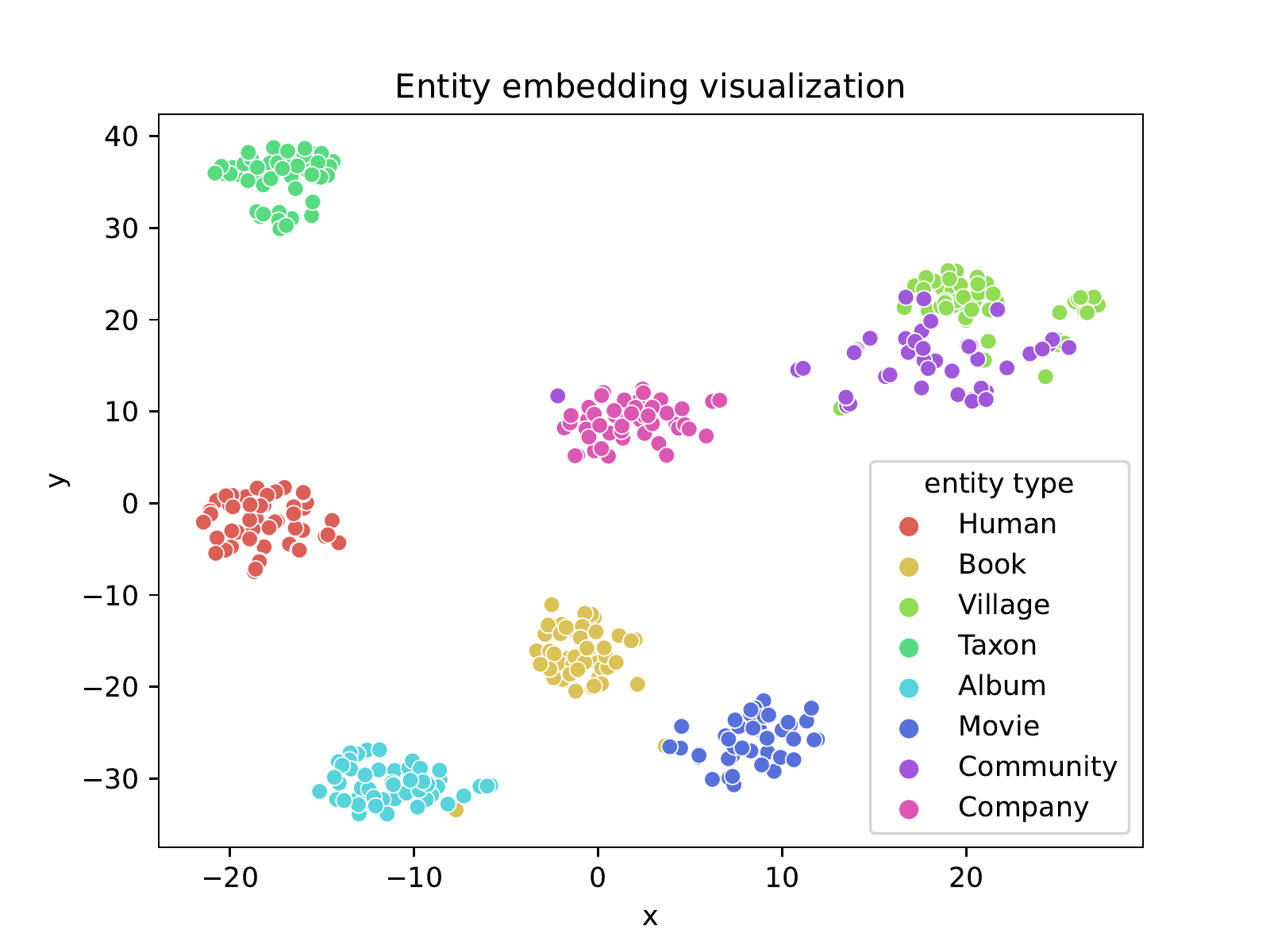}
 \caption{2-D visualization of the entity embeddings
 from the Wikidata5M-Trans dataset with t-SNE ~\citep{Maaten2008VisualizingDU}.}
 \label{fig:entity_vis}
\end{center}
\end{figure}

To examine our proposed model qualitatively,
we visualize the entity embeddings from $8$ largest categories
~\footnote{We utilize the ``instance of'' relation to determine the entity category.}
with $50$ randomly selected entities per category.
Entity embeddings are computed with BERT$_t$ in Section ~\ref{sec:arch}.
In Figure ~\ref{fig:entity_vis},
different categories are well separated,
demonstrating the high quality of the learned embeddings.
One interesting phenomenon is that
the two categories ``Community'' and ``Village'' have some overlap.
This is reasonable since these two concepts are not mutually exclusive.

\section{Conclusion}
This paper proposes a simple method SimKGC
to improve text-based knowledge graph completion.
We identify that the key issue is how to perform efficient contrastive learning.
Leveraging the recent progress in the field of contrastive learning,
SimKGC adopts a bi-encoder architecture
and combines three types of negatives.
Experiments on the WN18RR, FB15k-237, and Wikidata5M datasets
show that SimKGC substantially outperforms state-of-the-art methods.

For future work,
one direction is to improve the interpretability of SimKGC.
In methods like RotatE ~\citep{sun2018rotate} and TransE ~\citep{bordes2013translating},
a triple can be modeled as rotation in complex space or relational translation,
while SimKGC does not enable such easy-to-understand interpretations.
Another direction is to explore effective ways to deal with false negatives ~\citep{Huynh2020BoostingCS}
resulting from the incompleteness of knowledge graphs.

\section{Broader Impacts}
Future work could use SimKGC as a solid baseline
to keep improving text-based knowledge graph completion systems.
Our experimental results and analyses also reveal several promising research directions.
For example,
how to incorporate global graph structure in a more principled way?
Are there other loss functions that perform better than the InfoNCE loss?
For knowledge-intensive tasks such as knowledge base question answering (KBQA),
information retrieval, and knowledge-grounded response generation, etc.,
it would be interesting to explore the new opportunities
brought by the improved knowledge graph completion systems.

\section*{Acknowledgements}
We would like to thank anonymous reviewers and area chairs for their valuable comments,
and ACL Rolling Review organizers for their efforts.

\bibliography{anthology,custom}

\begin{thebibliography}{59}
\expandafter\ifx\csname natexlab\endcsname\relax\def\natexlab#1{#1}\fi

\bibitem[{Balazevic et~al.(2019)Balazevic, Allen, and
  Hospedales}]{balazevic-etal-2019-tucker}
Ivana Balazevic, Carl Allen, and Timothy Hospedales. 2019.
\newblock \href {https://doi.org/10.18653/v1/D19-1522} {{T}uck{ER}: Tensor
  factorization for knowledge graph completion}.
\newblock In \emph{Proceedings of the 2019 Conference on Empirical Methods in
  Natural Language Processing and the 9th International Joint Conference on
  Natural Language Processing (EMNLP-IJCNLP)}, pages 5185--5194, Hong Kong,
  China. Association for Computational Linguistics.

\bibitem[{Bollacker et~al.(2008)Bollacker, Evans, Paritosh, Sturge, and
  Taylor}]{bollacker2008freebase}
Kurt Bollacker, Colin Evans, Praveen Paritosh, Tim Sturge, and Jamie Taylor.
  2008.
\newblock Freebase: a collaboratively created graph database for structuring
  human knowledge.
\newblock In \emph{Proceedings of the 2008 ACM SIGMOD international conference
  on Management of data}, pages 1247--1250.

\bibitem[{Bordes et~al.(2013)Bordes, Usunier, Garc{\'{\i}}a{-}Dur{\'{a}}n,
  Weston, and Yakhnenko}]{bordes2013translating}
Antoine Bordes, Nicolas Usunier, Alberto Garc{\'{\i}}a{-}Dur{\'{a}}n, Jason
  Weston, and Oksana Yakhnenko. 2013.
\newblock \href
  {https://proceedings.neurips.cc/paper/2013/hash/1cecc7a77928ca8133fa24680a88d2f9-Abstract.html}
  {Translating embeddings for modeling multi-relational data}.
\newblock In \emph{Advances in Neural Information Processing Systems 26: 27th
  Annual Conference on Neural Information Processing Systems 2013. Proceedings
  of a meeting held December 5-8, 2013, Lake Tahoe, Nevada, United States},
  pages 2787--2795.

\bibitem[{Brown et~al.(2020)Brown, Mann, Ryder, Subbiah, Kaplan, Dhariwal,
  Neelakantan, Shyam, Sastry, Askell, Agarwal, Herbert{-}Voss, Krueger,
  Henighan, Child, Ramesh, Ziegler, Wu, Winter, Hesse, Chen, Sigler, Litwin,
  Gray, Chess, Clark, Berner, McCandlish, Radford, Sutskever, and
  Amodei}]{Brown2020LanguageMA}
Tom~B. Brown, Benjamin Mann, Nick Ryder, Melanie Subbiah, Jared Kaplan,
  Prafulla Dhariwal, Arvind Neelakantan, Pranav Shyam, Girish Sastry, Amanda
  Askell, Sandhini Agarwal, Ariel Herbert{-}Voss, Gretchen Krueger, Tom
  Henighan, Rewon Child, Aditya Ramesh, Daniel~M. Ziegler, Jeffrey Wu, Clemens
  Winter, Christopher Hesse, Mark Chen, Eric Sigler, Mateusz Litwin, Scott
  Gray, Benjamin Chess, Jack Clark, Christopher Berner, Sam McCandlish, Alec
  Radford, Ilya Sutskever, and Dario Amodei. 2020.
\newblock \href
  {https://proceedings.neurips.cc/paper/2020/hash/1457c0d6bfcb4967418bfb8ac142f64a-Abstract.html}
  {Language models are few-shot learners}.
\newblock In \emph{Advances in Neural Information Processing Systems 33: Annual
  Conference on Neural Information Processing Systems 2020, NeurIPS 2020,
  December 6-12, 2020, virtual}.

\bibitem[{Cao et~al.(2021)Cao, Ji, Lv, Li, Wen, and
  Zhang}]{cao-etal-2021-missing}
Yixin Cao, Xiang Ji, Xin Lv, Juanzi Li, Yonggang Wen, and Hanwang Zhang. 2021.
\newblock \href {https://doi.org/10.18653/v1/2021.acl-long.534} {Are missing
  links predictable? an inferential benchmark for knowledge graph completion}.
\newblock In \emph{Proceedings of the 59th Annual Meeting of the Association
  for Computational Linguistics and the 11th International Joint Conference on
  Natural Language Processing (Volume 1: Long Papers)}, pages 6855--6865,
  Online. Association for Computational Linguistics.

\bibitem[{Chen et~al.(2020)Chen, Kornblith, Norouzi, and
  Hinton}]{chen2020simple}
Ting Chen, Simon Kornblith, Mohammad Norouzi, and Geoffrey~E. Hinton. 2020.
\newblock \href {http://proceedings.mlr.press/v119/chen20j.html} {A simple
  framework for contrastive learning of visual representations}.
\newblock In \emph{Proceedings of the 37th International Conference on Machine
  Learning, {ICML} 2020, 13-18 July 2020, Virtual Event}, volume 119 of
  \emph{Proceedings of Machine Learning Research}, pages 1597--1607. {PMLR}.

\bibitem[{Daza et~al.(2021)Daza, Cochez, and Groth}]{daza2021inductive}
Daniel Daza, Michael Cochez, and Paul Groth. 2021.
\newblock Inductive entity representations from text via link prediction.
\newblock In \emph{Proceedings of the Web Conference 2021}, pages 798--808.

\bibitem[{Dettmers et~al.(2018)Dettmers, Minervini, Stenetorp, and
  Riedel}]{dettmers2018convolutional}
Tim Dettmers, Pasquale Minervini, Pontus Stenetorp, and Sebastian Riedel. 2018.
\newblock \href
  {https://www.aaai.org/ocs/index.php/AAAI/AAAI18/paper/view/17366}
  {Convolutional 2d knowledge graph embeddings}.
\newblock In \emph{Proceedings of the Thirty-Second {AAAI} Conference on
  Artificial Intelligence, (AAAI-18), the 30th innovative Applications of
  Artificial Intelligence (IAAI-18), and the 8th {AAAI} Symposium on
  Educational Advances in Artificial Intelligence (EAAI-18), New Orleans,
  Louisiana, USA, February 2-7, 2018}, pages 1811--1818. {AAAI} Press.

\bibitem[{Devlin et~al.(2019)Devlin, Chang, Lee, and
  Toutanova}]{devlin-etal-2019-bert}
Jacob Devlin, Ming-Wei Chang, Kenton Lee, and Kristina Toutanova. 2019.
\newblock \href {https://doi.org/10.18653/v1/N19-1423} {{BERT}: Pre-training of
  deep bidirectional transformers for language understanding}.
\newblock In \emph{Proceedings of the 2019 Conference of the North {A}merican
  Chapter of the Association for Computational Linguistics: Human Language
  Technologies, Volume 1 (Long and Short Papers)}, pages 4171--4186,
  Minneapolis, Minnesota. Association for Computational Linguistics.

\bibitem[{Dinan et~al.(2019)Dinan, Roller, Shuster, Fan, Auli, and
  Weston}]{Dinan2019WizardOW}
Emily Dinan, Stephen Roller, Kurt Shuster, Angela Fan, Michael Auli, and Jason
  Weston. 2019.
\newblock \href {https://openreview.net/forum?id=r1l73iRqKm} {Wizard of
  wikipedia: Knowledge-powered conversational agents}.
\newblock In \emph{7th International Conference on Learning Representations,
  {ICLR} 2019, New Orleans, LA, USA, May 6-9, 2019}. OpenReview.net.

\bibitem[{Gao et~al.(2021)Gao, Yao, and Chen}]{gao2021simcse}
Tianyu Gao, Xingcheng Yao, and Danqi Chen. 2021.
\newblock \href {https://arxiv.org/abs/2104.08821} {Simcse: Simple contrastive
  learning of sentence embeddings}.
\newblock \emph{ArXiv preprint}, abs/2104.08821.

\bibitem[{Grill et~al.(2020)Grill, Strub, Altch{\'{e}}, Tallec, Richemond,
  Buchatskaya, Doersch, Pires, Guo, Azar, Piot, Kavukcuoglu, Munos, and
  Valko}]{grill2020bootstrap}
Jean{-}Bastien Grill, Florian Strub, Florent Altch{\'{e}}, Corentin Tallec,
  Pierre~H. Richemond, Elena Buchatskaya, Carl Doersch, Bernardo~{\'{A}}vila
  Pires, Zhaohan Guo, Mohammad~Gheshlaghi Azar, Bilal Piot, Koray Kavukcuoglu,
  R{\'{e}}mi Munos, and Michal Valko. 2020.
\newblock \href
  {https://proceedings.neurips.cc/paper/2020/hash/f3ada80d5c4ee70142b17b8192b2958e-Abstract.html}
  {Bootstrap your own latent - {A} new approach to self-supervised learning}.
\newblock In \emph{Advances in Neural Information Processing Systems 33: Annual
  Conference on Neural Information Processing Systems 2020, NeurIPS 2020,
  December 6-12, 2020, virtual}.

\bibitem[{Gunel et~al.(2021)Gunel, Du, Conneau, and
  Stoyanov}]{gunel2020supervised}
Beliz Gunel, Jingfei Du, Alexis Conneau, and Veselin Stoyanov. 2021.
\newblock \href {https://openreview.net/forum?id=cu7IUiOhujH} {Supervised
  contrastive learning for pre-trained language model fine-tuning}.
\newblock In \emph{9th International Conference on Learning Representations,
  {ICLR} 2021, Virtual Event, Austria, May 3-7, 2021}. OpenReview.net.

\bibitem[{He et~al.(2020)He, Fan, Wu, Xie, and Girshick}]{he2020momentum}
Kaiming He, Haoqi Fan, Yuxin Wu, Saining Xie, and Ross~B. Girshick. 2020.
\newblock \href {https://doi.org/10.1109/CVPR42600.2020.00975} {Momentum
  contrast for unsupervised visual representation learning}.
\newblock In \emph{2020 {IEEE/CVF} Conference on Computer Vision and Pattern
  Recognition, {CVPR} 2020, Seattle, WA, USA, June 13-19, 2020}, pages
  9726--9735. {IEEE}.

\bibitem[{Huang et~al.(2018)Huang, Zhao, Dou, Wen, and
  Chang}]{Huang2018ImprovingSR}
Jin Huang, Wayne~Xin Zhao, Hongjian Dou, Ji{-}Rong Wen, and Edward~Y. Chang.
  2018.
\newblock \href {https://doi.org/10.1145/3209978.3210017} {Improving sequential
  recommendation with knowledge-enhanced memory networks}.
\newblock In \emph{The 41st International {ACM} {SIGIR} Conference on Research
  {\&} Development in Information Retrieval, {SIGIR} 2018, Ann Arbor, MI, USA,
  July 08-12, 2018}, pages 505--514. {ACM}.

\bibitem[{Huynh et~al.(2020)Huynh, Kornblith, Walter, Maire, and
  Khademi}]{Huynh2020BoostingCS}
Tri Huynh, Simon Kornblith, Matthew~R. Walter, Michael Maire, and Maryam
  Khademi. 2020.
\newblock \href {https://arxiv.org/abs/2011.11765} {Boosting contrastive
  self-supervised learning with false negative cancellation}.
\newblock \emph{ArXiv preprint}, abs/2011.11765.

\bibitem[{Jiang et~al.(2020)Jiang, Xu, Araki, and
  Neubig}]{jiang-etal-2020-know}
Zhengbao Jiang, Frank~F. Xu, Jun Araki, and Graham Neubig. 2020.
\newblock \href {https://doi.org/10.1162/tacl_a_00324} {How can we know what
  language models know?}
\newblock \emph{Transactions of the Association for Computational Linguistics},
  8:423--438.

\bibitem[{Johnson et~al.(2021)Johnson, Douze, and
  J{\'e}gou}]{Johnson2021BillionScaleSS}
Jeff Johnson, M.~Douze, and H.~J{\'e}gou. 2021.
\newblock Billion-scale similarity search with gpus.
\newblock \emph{IEEE Transactions on Big Data}, 7:535--547.

\bibitem[{Karpukhin et~al.(2020)Karpukhin, Oguz, Min, Lewis, Wu, Edunov, Chen,
  and Yih}]{karpukhin-etal-2020-dense}
Vladimir Karpukhin, Barlas Oguz, Sewon Min, Patrick Lewis, Ledell Wu, Sergey
  Edunov, Danqi Chen, and Wen-tau Yih. 2020.
\newblock \href {https://doi.org/10.18653/v1/2020.emnlp-main.550} {Dense
  passage retrieval for open-domain question answering}.
\newblock In \emph{Proceedings of the 2020 Conference on Empirical Methods in
  Natural Language Processing (EMNLP)}, pages 6769--6781, Online. Association
  for Computational Linguistics.

\bibitem[{Kim et~al.(2020)Kim, Hong, Ko, and Seo}]{kim-etal-2020-multi}
Bosung Kim, Taesuk Hong, Youngjoong Ko, and Jungyun Seo. 2020.
\newblock \href {https://doi.org/10.18653/v1/2020.coling-main.153} {Multi-task
  learning for knowledge graph completion with pre-trained language models}.
\newblock In \emph{Proceedings of the 28th International Conference on
  Computational Linguistics}, pages 1737--1743, Barcelona, Spain (Online).
  International Committee on Computational Linguistics.

\bibitem[{Le-Khac et~al.(2020)Le-Khac, Healy, and Smeaton}]{le2020contrastive}
Phuc~H Le-Khac, Graham Healy, and Alan~F Smeaton. 2020.
\newblock Contrastive representation learning: A framework and review.
\newblock \emph{IEEE Access}.

\bibitem[{Lee et~al.(2021)Lee, Sung, Kang, and Chen}]{Lee2021LearningDR}
Jinhyuk Lee, Mujeen Sung, Jaewoo Kang, and Danqi Chen. 2021.
\newblock \href {https://doi.org/10.18653/v1/2021.acl-long.518} {Learning dense
  representations of phrases at scale}.
\newblock In \emph{Proceedings of the 59th Annual Meeting of the Association
  for Computational Linguistics and the 11th International Joint Conference on
  Natural Language Processing (Volume 1: Long Papers)}, pages 6634--6647,
  Online. Association for Computational Linguistics.

\bibitem[{Liu et~al.(2021)Liu, Fan, Qian, Chen, Ding, and Wang}]{Liu2021HiTHT}
Song Liu, Haoqi Fan, Shengsheng Qian, Yiru Chen, W.~Ding, and Zhongyuan Wang.
  2021.
\newblock \href {https://arxiv.org/abs/2103.15049} {Hit: Hierarchical
  transformer with momentum contrast for video-text retrieval}.
\newblock \emph{ArXiv preprint}, abs/2103.15049.

\bibitem[{Liu et~al.(2020)Liu, Zhou, Zhao, Wang, Ju, Deng, and Wang}]{liu2020k}
Weijie Liu, Peng Zhou, Zhe Zhao, Zhiruo Wang, Qi~Ju, Haotang Deng, and Ping
  Wang. 2020.
\newblock \href {https://aaai.org/ojs/index.php/AAAI/article/view/5681}
  {{K-BERT:} enabling language representation with knowledge graph}.
\newblock In \emph{The Thirty-Fourth {AAAI} Conference on Artificial
  Intelligence, {AAAI} 2020, The Thirty-Second Innovative Applications of
  Artificial Intelligence Conference, {IAAI} 2020, The Tenth {AAAI} Symposium
  on Educational Advances in Artificial Intelligence, {EAAI} 2020, New York,
  NY, USA, February 7-12, 2020}, pages 2901--2908. {AAAI} Press.

\bibitem[{Maaten and Hinton(2008)}]{Maaten2008VisualizingDU}
L.~V.~D. Maaten and Geoffrey~E. Hinton. 2008.
\newblock Visualizing data using t-sne.
\newblock \emph{Journal of Machine Learning Research}, 9:2579--2605.

\bibitem[{Malaviya et~al.(2020)Malaviya, Bhagavatula, Bosselut, and
  Choi}]{Malaviya2020CommonsenseKB}
Chaitanya Malaviya, Chandra Bhagavatula, Antoine Bosselut, and Yejin Choi.
  2020.
\newblock Commonsense knowledge base completion with structural and semantic
  context.
\newblock In \emph{AAAI}.

\bibitem[{Miller(1992)}]{miller1995wordnet}
George~A. Miller. 1992.
\newblock \href {https://aclanthology.org/H92-1116} {{W}ord{N}et: A lexical
  database for {E}nglish}.
\newblock In \emph{Speech and Natural Language: Proceedings of a Workshop Held
  at Harriman, New York, {F}ebruary 23-26, 1992}.

\bibitem[{Ni et~al.(2021)Ni, Constant, Ma, Hall, Cer, Yang
  et~al.}]{ni2021sentence}
Jianmo Ni, Noah Constant, Ji~Ma, Keith~B Hall, Daniel Cer, Yinfei Yang, et~al.
  2021.
\newblock \href {https://arxiv.org/abs/2108.08877} {Sentence-t5: Scalable
  sentence encoders from pre-trained text-to-text models}.
\newblock \emph{ArXiv preprint}, abs/2108.08877.

\bibitem[{Nickel et~al.(2011)Nickel, Tresp, and Kriegel}]{nickel2011three}
Maximilian Nickel, Volker Tresp, and Hans{-}Peter Kriegel. 2011.
\newblock \href {https://icml.cc/2011/papers/438\_icmlpaper.pdf} {A three-way
  model for collective learning on multi-relational data}.
\newblock In \emph{Proceedings of the 28th International Conference on Machine
  Learning, {ICML} 2011, Bellevue, Washington, USA, June 28 - July 2, 2011},
  pages 809--816. Omnipress.

\bibitem[{Petroni et~al.(2019)Petroni, Rockt{\"a}schel, Riedel, Lewis, Bakhtin,
  Wu, and Miller}]{petroni-etal-2019-language}
Fabio Petroni, Tim Rockt{\"a}schel, Sebastian Riedel, Patrick Lewis, Anton
  Bakhtin, Yuxiang Wu, and Alexander Miller. 2019.
\newblock \href {https://doi.org/10.18653/v1/D19-1250} {Language models as
  knowledge bases?}
\newblock In \emph{Proceedings of the 2019 Conference on Empirical Methods in
  Natural Language Processing and the 9th International Joint Conference on
  Natural Language Processing (EMNLP-IJCNLP)}, pages 2463--2473, Hong Kong,
  China. Association for Computational Linguistics.

\bibitem[{Qu et~al.(2021)Qu, Ding, Liu, Liu, Ren, Zhao, Dong, Wu, and
  Wang}]{qu-etal-2021-rocketqa}
Yingqi Qu, Yuchen Ding, Jing Liu, Kai Liu, Ruiyang Ren, Wayne~Xin Zhao, Daxiang
  Dong, Hua Wu, and Haifeng Wang. 2021.
\newblock \href {https://doi.org/10.18653/v1/2021.naacl-main.466}
  {{R}ocket{QA}: An optimized training approach to dense passage retrieval for
  open-domain question answering}.
\newblock In \emph{Proceedings of the 2021 Conference of the North American
  Chapter of the Association for Computational Linguistics: Human Language
  Technologies}, pages 5835--5847, Online. Association for Computational
  Linguistics.

\bibitem[{Radford et~al.(2021)Radford, Kim, Hallacy, Ramesh, Goh, Agarwal,
  Sastry, Askell, Mishkin, Clark, Krueger, and Sutskever}]{radford2021learning}
Alec Radford, Jong~Wook Kim, Chris Hallacy, Aditya Ramesh, Gabriel Goh,
  Sandhini Agarwal, Girish Sastry, Amanda Askell, Pamela Mishkin, Jack Clark,
  Gretchen Krueger, and Ilya Sutskever. 2021.
\newblock \href {http://proceedings.mlr.press/v139/radford21a.html} {Learning
  transferable visual models from natural language supervision}.
\newblock In \emph{Proceedings of the 38th International Conference on Machine
  Learning, {ICML} 2021, 18-24 July 2021, Virtual Event}, volume 139 of
  \emph{Proceedings of Machine Learning Research}, pages 8748--8763. {PMLR}.

\bibitem[{Radford et~al.(2018)Radford, Narasimhan, Salimans, and
  Sutskever}]{radford2018improving}
Alec Radford, Karthik Narasimhan, Tim Salimans, and Ilya Sutskever. 2018.
\newblock Improving language understanding by generative pre-training.

\bibitem[{Raffel et~al.(2019)Raffel, Shazeer, Roberts, Lee, Narang, Matena,
  Zhou, Li, and Liu}]{Raffel2020ExploringTL}
Colin Raffel, Noam~M. Shazeer, Adam Roberts, Katherine Lee, Sharan Narang,
  Michael Matena, Yanqi Zhou, W.~Li, and Peter~J. Liu. 2019.
\newblock \href {https://arxiv.org/abs/1910.10683} {Exploring the limits of
  transfer learning with a unified text-to-text transformer}.
\newblock \emph{ArXiv preprint}, abs/1910.10683.

\bibitem[{Reimers and Gurevych(2019)}]{Reimers2019SentenceBERTSE}
Nils Reimers and Iryna Gurevych. 2019.
\newblock \href {https://doi.org/10.18653/v1/D19-1410} {Sentence-{BERT}:
  Sentence embeddings using {S}iamese {BERT}-networks}.
\newblock In \emph{Proceedings of the 2019 Conference on Empirical Methods in
  Natural Language Processing and the 9th International Joint Conference on
  Natural Language Processing (EMNLP-IJCNLP)}, pages 3982--3992, Hong Kong,
  China. Association for Computational Linguistics.

\bibitem[{Shin et~al.(2020)Shin, Razeghi, Logan~IV, Wallace, and
  Singh}]{shin-etal-2020-autoprompt}
Taylor Shin, Yasaman Razeghi, Robert~L. Logan~IV, Eric Wallace, and Sameer
  Singh. 2020.
\newblock \href {https://doi.org/10.18653/v1/2020.emnlp-main.346}
  {{A}uto{P}rompt: {E}liciting {K}nowledge from {L}anguage {M}odels with
  {A}utomatically {G}enerated {P}rompts}.
\newblock In \emph{Proceedings of the 2020 Conference on Empirical Methods in
  Natural Language Processing (EMNLP)}, pages 4222--4235, Online. Association
  for Computational Linguistics.

\bibitem[{Speer et~al.(2017)Speer, Chin, and Havasi}]{speer2017conceptnet}
Robyn Speer, Joshua Chin, and Catherine Havasi. 2017.
\newblock \href {http://aaai.org/ocs/index.php/AAAI/AAAI17/paper/view/14972}
  {Conceptnet 5.5: An open multilingual graph of general knowledge}.
\newblock In \emph{Proceedings of the Thirty-First {AAAI} Conference on
  Artificial Intelligence, February 4-9, 2017, San Francisco, California,
  {USA}}, pages 4444--4451. {AAAI} Press.

\bibitem[{Suchanek et~al.(2007)Suchanek, Kasneci, and
  Weikum}]{suchanek2007yago}
Fabian~M. Suchanek, Gjergji Kasneci, and Gerhard Weikum. 2007.
\newblock \href {https://doi.org/10.1145/1242572.1242667} {Yago: a core of
  semantic knowledge}.
\newblock In \emph{Proceedings of the 16th International Conference on World
  Wide Web, {WWW} 2007, Banff, Alberta, Canada, May 8-12, 2007}, pages
  697--706. {ACM}.

\bibitem[{Sun et~al.(2019{\natexlab{a}})Sun, Bedrax-Weiss, and
  Cohen}]{sun-etal-2019-pullnet}
Haitian Sun, Tania Bedrax-Weiss, and William Cohen. 2019{\natexlab{a}}.
\newblock \href {https://doi.org/10.18653/v1/D19-1242} {{P}ull{N}et: Open
  domain question answering with iterative retrieval on knowledge bases and
  text}.
\newblock In \emph{Proceedings of the 2019 Conference on Empirical Methods in
  Natural Language Processing and the 9th International Joint Conference on
  Natural Language Processing (EMNLP-IJCNLP)}, pages 2380--2390, Hong Kong,
  China. Association for Computational Linguistics.

\bibitem[{Sun et~al.(2019{\natexlab{b}})Sun, Deng, Nie, and
  Tang}]{sun2018rotate}
Zhiqing Sun, Zhi{-}Hong Deng, Jian{-}Yun Nie, and Jian Tang.
  2019{\natexlab{b}}.
\newblock \href {https://openreview.net/forum?id=HkgEQnRqYQ} {Rotate: Knowledge
  graph embedding by relational rotation in complex space}.
\newblock In \emph{7th International Conference on Learning Representations,
  {ICLR} 2019, New Orleans, LA, USA, May 6-9, 2019}. OpenReview.net.

\bibitem[{Teru et~al.(2020)Teru, Denis, and Hamilton}]{Teru2020InductiveRP}
Komal Teru, Etienne Denis, and Will Hamilton. 2020.
\newblock \href {http://proceedings.mlr.press/v119/teru20a.html} {Inductive
  relation prediction by subgraph reasoning}.
\newblock In \emph{Proceedings of the 37th International Conference on Machine
  Learning, {ICML} 2020, 13-18 July 2020, Virtual Event}, volume 119 of
  \emph{Proceedings of Machine Learning Research}, pages 9448--9457. {PMLR}.

\bibitem[{Toutanova et~al.(2015)Toutanova, Chen, Pantel, Poon, Choudhury, and
  Gamon}]{toutanova-etal-2015-representing}
Kristina Toutanova, Danqi Chen, Patrick Pantel, Hoifung Poon, Pallavi
  Choudhury, and Michael Gamon. 2015.
\newblock \href {https://doi.org/10.18653/v1/D15-1174} {Representing text for
  joint embedding of text and knowledge bases}.
\newblock In \emph{Proceedings of the 2015 Conference on Empirical Methods in
  Natural Language Processing}, pages 1499--1509, Lisbon, Portugal. Association
  for Computational Linguistics.

\bibitem[{Trouillon et~al.(2016)Trouillon, Welbl, Riedel, Gaussier, and
  Bouchard}]{trouillon2016complex}
Th{\'{e}}o Trouillon, Johannes Welbl, Sebastian Riedel, {\'{E}}ric Gaussier,
  and Guillaume Bouchard. 2016.
\newblock \href {http://proceedings.mlr.press/v48/trouillon16.html} {Complex
  embeddings for simple link prediction}.
\newblock In \emph{Proceedings of the 33nd International Conference on Machine
  Learning, {ICML} 2016, New York City, NY, USA, June 19-24, 2016}, volume~48
  of \emph{{JMLR} Workshop and Conference Proceedings}, pages 2071--2080.
  JMLR.org.

\bibitem[{Vrande{\v{c}}i{\'c} and Kr{\"o}tzsch(2014)}]{vrandevcic2014wikidata}
Denny Vrande{\v{c}}i{\'c} and Markus Kr{\"o}tzsch. 2014.
\newblock Wikidata: a free collaborative knowledgebase.
\newblock \emph{Communications of the ACM}, 57(10):78--85.

\bibitem[{Wang et~al.(2021{\natexlab{a}})Wang, Shen, Long, Zhou, Wang, and
  Chang}]{wang2021structure}
Bo~Wang, Tao Shen, Guodong Long, Tianyi Zhou, Ying Wang, and Yi~Chang.
  2021{\natexlab{a}}.
\newblock Structure-augmented text representation learning for efficient
  knowledge graph completion.
\newblock In \emph{Proceedings of the Web Conference 2021}, pages 1737--1748.

\bibitem[{Wang and Liu(2021)}]{Wang2021UnderstandingTB}
Feng Wang and Huaping Liu. 2021.
\newblock Understanding the behaviour of contrastive loss.
\newblock \emph{2021 IEEE/CVF Conference on Computer Vision and Pattern
  Recognition (CVPR)}, pages 2495--2504.

\bibitem[{Wang et~al.(2021{\natexlab{b}})Wang, Zhao, and
  Liu}]{Wang2021AligningCS}
Liang Wang, Wei Zhao, and Jingming Liu. 2021{\natexlab{b}}.
\newblock Aligning cross-lingual sentence representations with dual momentum
  contrast.
\newblock In \emph{EMNLP}.

\bibitem[{Wang et~al.(2021{\natexlab{c}})Wang, Gao, Zhu, Zhang, Liu, Li, and
  Tang}]{wang2021kepler}
Xiaozhi Wang, Tianyu Gao, Zhaocheng Zhu, Zhengyan Zhang, Zhiyuan Liu, Juanzi
  Li, and Jian Tang. 2021{\natexlab{c}}.
\newblock Kepler: A unified model for knowledge embedding and pre-trained
  language representation.
\newblock \emph{Transactions of the Association for Computational Linguistics},
  9:176--194.

\bibitem[{Wang et~al.(2014)Wang, Zhang, Feng, and Chen}]{wang2014knowledge}
Zhen Wang, Jianwen Zhang, Jianlin Feng, and Zheng Chen. 2014.
\newblock \href {http://www.aaai.org/ocs/index.php/AAAI/AAAI14/paper/view/8531}
  {Knowledge graph embedding by translating on hyperplanes}.
\newblock In \emph{Proceedings of the Twenty-Eighth {AAAI} Conference on
  Artificial Intelligence, July 27 -31, 2014, Qu{\'{e}}bec City, Qu{\'{e}}bec,
  Canada}, pages 1112--1119. {AAAI} Press.

\bibitem[{Wu et~al.(2019)Wu, Pan, Chen, Long, Zhang, and Yu}]{Wu2019ACS}
Zonghan Wu, Shirui Pan, Fengwen Chen, Guodong Long, C.~Zhang, and Philip~S. Yu.
  2019.
\newblock A comprehensive survey on graph neural networks.
\newblock \emph{IEEE Transactions on Neural Networks and Learning Systems},
  32:4--24.

\bibitem[{Xie et~al.(2016)Xie, Liu, Jia, Luan, and Sun}]{xie2016representation}
Ruobing Xie, Zhiyuan Liu, Jia Jia, Huanbo Luan, and Maosong Sun. 2016.
\newblock \href
  {http://www.aaai.org/ocs/index.php/AAAI/AAAI16/paper/view/12216}
  {Representation learning of knowledge graphs with entity descriptions}.
\newblock In \emph{Proceedings of the Thirtieth {AAAI} Conference on Artificial
  Intelligence, February 12-17, 2016, Phoenix, Arizona, {USA}}, pages
  2659--2665. {AAAI} Press.

\bibitem[{Xiong et~al.(2021)Xiong, Xiong, Li, Tang, Liu, Bennett, Ahmed, and
  Overwijk}]{xiong2020approximate}
Lee Xiong, Chenyan Xiong, Ye~Li, Kwok{-}Fung Tang, Jialin Liu, Paul~N. Bennett,
  Junaid Ahmed, and Arnold Overwijk. 2021.
\newblock \href {https://openreview.net/forum?id=zeFrfgyZln} {Approximate
  nearest neighbor negative contrastive learning for dense text retrieval}.
\newblock In \emph{9th International Conference on Learning Representations,
  {ICLR} 2021, Virtual Event, Austria, May 3-7, 2021}. OpenReview.net.

\bibitem[{Yang et~al.(2015)Yang, Yih, He, Gao, and Deng}]{yang2014embedding}
Bishan Yang, Wen{-}tau Yih, Xiaodong He, Jianfeng Gao, and Li~Deng. 2015.
\newblock \href {http://arxiv.org/abs/1412.6575} {Embedding entities and
  relations for learning and inference in knowledge bases}.
\newblock In \emph{3rd International Conference on Learning Representations,
  {ICLR} 2015, San Diego, CA, USA, May 7-9, 2015, Conference Track
  Proceedings}.

\bibitem[{Yang et~al.(2019)Yang, {\'{A}}brego, Yuan, Guo, Shen, Cer, Sung,
  Strope, and Kurzweil}]{yang2019improving}
Yinfei Yang, Gustavo~Hern{\'{a}}ndez {\'{A}}brego, Steve Yuan, Mandy Guo,
  Qinlan Shen, Daniel Cer, Yun{-}Hsuan Sung, Brian Strope, and Ray Kurzweil.
  2019.
\newblock \href {https://doi.org/10.24963/ijcai.2019/746} {Improving
  multilingual sentence embedding using bi-directional dual encoder with
  additive margin softmax}.
\newblock In \emph{Proceedings of the Twenty-Eighth International Joint
  Conference on Artificial Intelligence, {IJCAI} 2019, Macao, China, August
  10-16, 2019}, pages 5370--5378. ijcai.org.

\bibitem[{Yao et~al.(2019)Yao, Mao, and Luo}]{yao2019kg}
Liang Yao, Chengsheng Mao, and Yuan Luo. 2019.
\newblock \href {https://arxiv.org/abs/1909.03193} {Kg-bert: Bert for knowledge
  graph completion}.
\newblock \emph{ArXiv preprint}, abs/1909.03193.

\bibitem[{You et~al.(2020)You, Li, Reddi, Hseu, Kumar, Bhojanapalli, Song,
  Demmel, Keutzer, and Hsieh}]{You2020LargeBO}
Yang You, Jing Li, Sashank~J. Reddi, Jonathan Hseu, Sanjiv Kumar, Srinadh
  Bhojanapalli, Xiaodan Song, James Demmel, Kurt Keutzer, and Cho{-}Jui Hsieh.
  2020.
\newblock \href {https://openreview.net/forum?id=Syx4wnEtvH} {Large batch
  optimization for deep learning: Training {BERT} in 76 minutes}.
\newblock In \emph{8th International Conference on Learning Representations,
  {ICLR} 2020, Addis Ababa, Ethiopia, April 26-30, 2020}. OpenReview.net.

\bibitem[{Zha et~al.(2021)Zha, Chen, and Yan}]{Zha2021InductiveRP}
Hanwen Zha, Zhiyu Chen, and Xifeng Yan. 2021.
\newblock \href {https://arxiv.org/abs/2103.07102} {Inductive relation
  prediction by bert}.
\newblock \emph{ArXiv preprint}, abs/2103.07102.

\bibitem[{Zhang et~al.(2019)Zhang, Han, Liu, Jiang, Sun, and
  Liu}]{zhang-etal-2019-ernie}
Zhengyan Zhang, Xu~Han, Zhiyuan Liu, Xin Jiang, Maosong Sun, and Qun Liu. 2019.
\newblock \href {https://doi.org/10.18653/v1/P19-1139} {{ERNIE}: Enhanced
  language representation with informative entities}.
\newblock In \emph{Proceedings of the 57th Annual Meeting of the Association
  for Computational Linguistics}, pages 1441--1451, Florence, Italy.
  Association for Computational Linguistics.

\bibitem[{Zhong et~al.(2021)Zhong, Friedman, and Chen}]{zhong2021factual}
Zexuan Zhong, Dan Friedman, and Danqi Chen. 2021.
\newblock \href {https://doi.org/10.18653/v1/2021.naacl-main.398} {Factual
  probing is [{MASK}]: Learning vs. learning to recall}.
\newblock In \emph{Proceedings of the 2021 Conference of the North American
  Chapter of the Association for Computational Linguistics: Human Language
  Technologies}, pages 5017--5033, Online. Association for Computational
  Linguistics.

\end{thebibliography}
\bibliographystyle{acl_natbib}

\appendix

\section{Details on Hyperparameters} ~\label{app:setup}

\begin{table}[ht]
\centering
\scalebox{0.9}{\begin{tabular}{l|c}
\hline
Hyperparameter & value \\ \hline
\# of GPUs  &  4 \\ \hline
initial temperature $\tau$  &   0.05 \\ \hline
gradient clip & 10  \\ \hline
warmup steps & 400  \\ \hline
batch size  & 1024 \\ \hline
max \# of tokens  & 50 \\ \hline
weight $\alpha$ for re-ranking  & 0.05 \\ \hline
dropout & 0.1 \\ \hline
weight decay  & $10^{-4}$ \\ \hline
InfoNCE margin  & 0.02 \\ \hline
pooling  &  mean \\ \hline
\end{tabular}}
\caption{Shared hyperparameters for our proposed SimKGC model.}
\label{tab:appendix_hyper}
\end{table}

In Table ~\ref{tab:appendix_hyper},
we show the hyperparameters that are shared across all the datasets.
For learning rate,
we use $5\times10^{-5}$, $10^{-5}$, and $3\times10^{-5}$
for WN18RR, FB15k-237, and Wikidata5M datasets, respectively.
For re-ranking,
we use $5$-hop neighbors for WN18RR and $2$-hop neighbors for other datasets.
Each epoch takes $\sim3$ minutes for WN18RR,
$\sim12$ minutes for FB15k-237,
and $\sim12$ hours for Wikidata5M (both settings).
Our implementation is based on open-source project \emph{transformers}
~\footnote{\url{https://github.com/huggingface/transformers}}.

For inverse relation $r^{-1}$,
we add a prefix word ``inverse'' to the description of $r$.
For examples,
if $r$ = ``instance of'',
then $r^{-1}$ = ``inverse instance of''.

Some entities in the WN18RR and FB15k-237 dataset
have very short textual descriptions.
We concatenate them with the entity names of its neighbors in the training set.
To avoid label leakage during training,
we dynamically exclude the correct entity in the input text.

\section{More Analysis Results} \label{app:analysis}

\begin{table}[ht]
\centering
\begin{tabular}{c|llll}
\hline
batch size & MRR & H@1 & H@3 & H@10 \\ \hline
256 &  33.8 & 28.7 & 35.8 & 43.1 \\
512 & 34.6 & 29.4 & 36.7 & 43.7 \\
1024 & \textbf{35.3} & \textbf{30.1} & \textbf{37.4} & \textbf{44.8} \\ \hline
\end{tabular}
\caption{Effects of batch size on the Wikidata5M-Trans dataset with SimKGC$_\text{IB}$.}
\label{tab:wiki5m_trans_batch_size}
\end{table}

\begin{table}[ht]
\centering
\begin{tabular}{c|llll}
\hline
batch size & MRR & H@1 & H@3 & H@10 \\ \hline
256 &  32.4 & 23.3 & 35.4 & 50.9 \\
512 & 32.7 & 23.7 & 35.6 & \textbf{51.0} \\
1024 & \textbf{33.3} & \textbf{24.6} & \textbf{36.2} & \textbf{51.0} \\ \hline
\end{tabular}
\caption{Effects of batch size on the FB15k-237 dataset with SimKGC$_\text{IB}$.}
\label{tab:fb_batch_size}
\end{table}

\begin{table}[ht]
\centering
\scalebox{0.9}{\begin{tabular}{l|cccc}
\hline
margin $\gamma$ & MRR & H@1 & H@3 & H@10 \\ \hline
0 & 33.4 & 24.8 & 36.0 & 50.9 \\
0.02 & \textbf{33.6} & 24.9 & \textbf{36.2} & \textbf{51.1} \\
0.05 & \textbf{33.6} & \textbf{25.0} & \textbf{36.2} & 50.9 \\ \hline
\end{tabular}}
\caption{Ablation for the additive margin $\gamma$ of InfoNCE loss on the FB15k-237 dataset.}
\label{tab:fb_additive_margin}
\end{table}

\begin{table*}[ht]
\centering
\scalebox{0.9}{\begin{tabular}{l|l}
\hline
\multirow{2}{*}{triple} & \multirow{2}{*}{(captive state (film), instance of, \textbf{movie})} \\
 &   \\
evidence & \begin{tabular}[c]{@{}l@{}}Captive State is a 2019 American crime science fiction thriller film directed by Rupert Wyatt \\ and co-written by Wyatt and Erica Beeney.\ldots \end{tabular} \\
SimKGC &  3-D movies \\ \hline
triple   & (Lionel Belasco, occupation, \textbf{composer}) \\
evidence & \begin{tabular}[c]{@{}l@{}} Lionel Belasco (1881 – c. 24 June 1967) was a prominent pianist, composer and bandleader, \\ best known for his calypso recordings. \end{tabular} \\
SimKGC &  bandleaders \\ \hline
triple   & (\textbf{Johan Nordhagen}, country of citizenship, Norway) \\
evidence & Waqas Ahmed (born 9 June 1991) is a Norwegian cricketer. \ldots \\
SimKGC &  Waqas Ahmed  \\ \hline
triple   & (\textbf{Carlos Peña Romulo}, position held, philippine resident commissioner) \\
evidence & \begin{tabular}[c]{@{}l@{}} Francis Burton Harrison was an American-born Filipino statesman who served in the United States \\ House of Representatives and was appointed Governor-General of the Philippines \ldots \end{tabular} \\
SimKGC &  Francis Burton Harrison \\ \hline
\end{tabular}}
\caption{More examples of SimKGC prediction results on the test set of Wikidata5M-Trans.}
\label{tab:more_cases}
\end{table*}

In Table ~\ref{tab:wiki5m_trans_batch_size} and ~\ref{tab:fb_batch_size},
we show how the batch size affects model performance
on the Wikidata5M-Trans and FB15k-237 dataset.

In Equation ~\ref{eq:infonce},
we use a variant of InfoNCE loss that has an additive margin $\gamma$.
In our experiments,
such a variant performs consistently better than the standard InfoNCE loss,
though the improvement is quite marginal,
as shown in Table ~\ref{tab:fb_additive_margin}.

In Table ~\ref{tab:more_cases},
we show more examples of SimKGC predictions on the Wikidata5M-Trans dataset
to help better understand our model's behavior.
Full model predictions on test datasets are available in our public code repository.

\end{document}